\documentclass[11pt]{article}

\usepackage[final]{acl}

\usepackage{times}
\usepackage{latexsym}
\usepackage{amsmath}
\usepackage{enumitem}
\usepackage{booktabs} 

\usepackage{multirow}

\usepackage[table]{xcolor} 

\usepackage[T1]{fontenc}
\usepackage[utf8]{inputenc}

\usepackage[utf8]{inputenc}

\usepackage{microtype}

\usepackage{inconsolata}

\usepackage{graphicx}

\title{\textbf{\texttt{IS-CoT}}: Breaking the Long-form Generation Collapse \\ via Interleaved Structural Thinking}

\author{
  \textbf{Zechen Sun\textsuperscript{1}},
  \textbf{Yuyang Sun\textsuperscript{1}},
  \textbf{Zecheng Tang\textsuperscript{1}},
    \textbf{Juntao Li\textsuperscript{1}\thanks{Corresponding Authors}},
    \textbf{Wenpeng Hu\textsuperscript{2}\footnotemark[\value{footnote}]},
\\
  \textbf{Wenliang Chen\textsuperscript{1}},
  \textbf{Zhunchen Luo\textsuperscript{2}},
  \textbf{Guotong Geng\textsuperscript{2}},
  \textbf{Min Zhang\textsuperscript{1}}
\\
\\
  \textsuperscript{1}Institute of Computer Science and Technology, Soochow University \\
  \textsuperscript{2}Information Research Center of Military Science, PLA Academy of Military Science
\\
  zcsunc@stu.suda.edu.cn, ljt@suda.edu.cn, wenpeng.hu@pku.edu.cn \\
}

\begin{document}
\maketitle
\begin{abstract}
Generating coherent and controllable long-form content remains a persistent challenge for Large Language Models (LLMs). 
While reasoning-enhanced models have demonstrated success in logic-intensive domains, our evaluation reveals that they suffer from a severe length collapse in open-ended writing, where performance degrades sharply as target lengths exceed 2,000 words.
We attribute this failure to the limitation of static hierarchical planning, which struggles to provide dynamic guidance over extended contexts.
To bridge this gap, we introduce the \textbf{Interleaved Structural Chain-of-Thought (\texttt{IS-CoT})} framework.
Unlike external agentic workflows, \texttt{IS-CoT} embeds a dynamic \texttt{Plan-Write-Reflect} cycle into the generation process, enabling continuous strategy adaptation and global alignment without additional assistance.
Based on this framework, we construct a high-quality dataset of interleaved reasoning traces via a multi-teacher pipeline and train \textbf{IS-Writer-8B}. Experiments demonstrate that IS-Writer-8B achieves state-of-the-art performance on challenging long-form benchmarks~(e.g., +3.08 vs. DeepSeek-V3.2 on LongBench-Write), exhibiting robust length compliance and coherence competitive with significantly larger proprietary models.
\end{abstract}


\section{Introduction}
\begin{figure}[h] 
\centering
\includegraphics[scale=0.11]{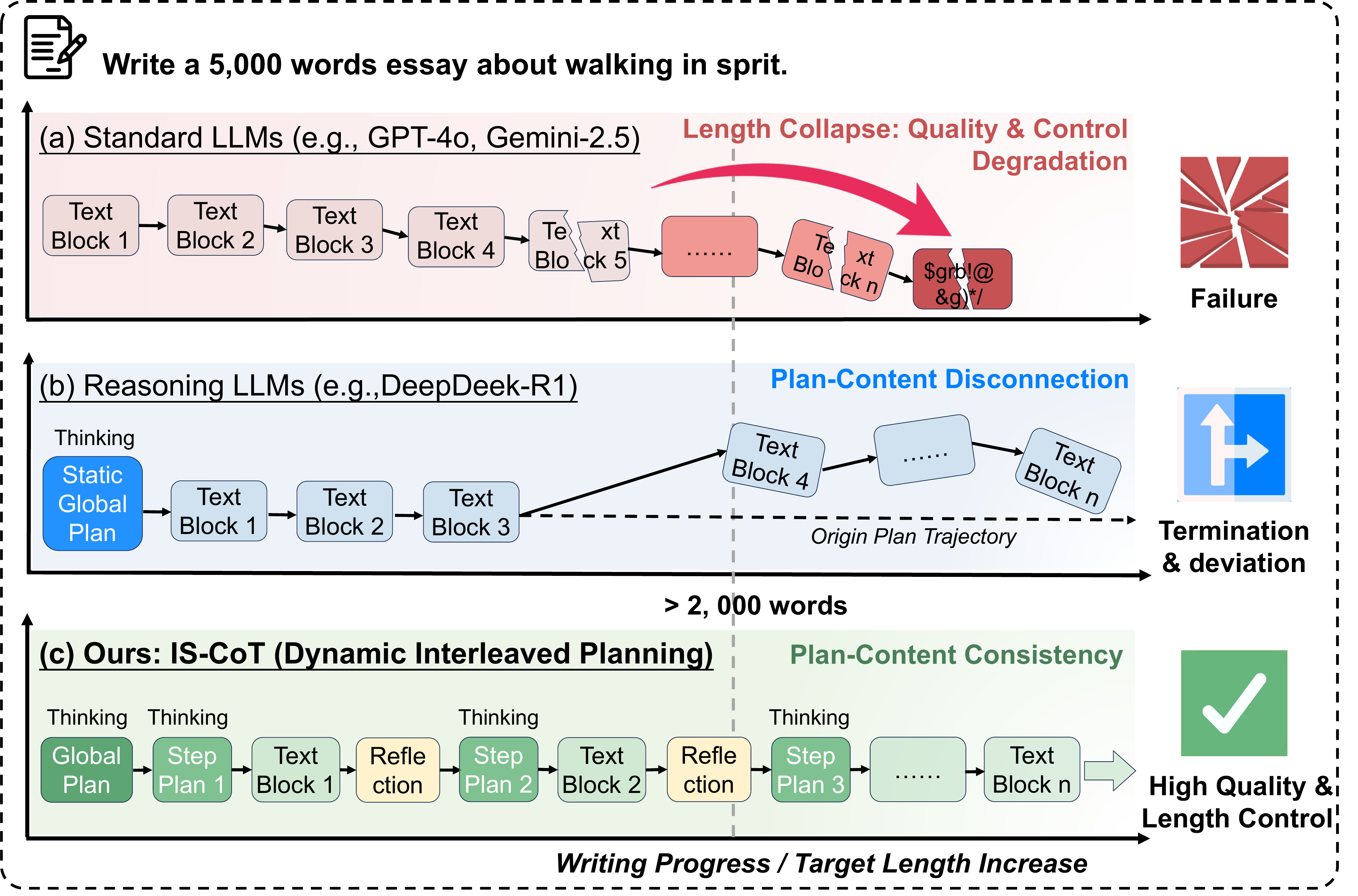}
\caption{Comparison of three long-form generation paradigms. While existing methods degrade as target length increases due to static planning, our \texttt{IS-CoT} introduces a dynamic Plan-Write-Reflect cycle, maintaining coherence and length control over long horizons.}
\label{fig:paradigm}
\end{figure}

With the remarkable success of large language models (LLMs) across various generation tasks~\cite{comanici2025gemini,deepseekai2025deepseekv32pushingfrontieropen,anthropic_claude35_sonnet_2024}, there is a growing demand for generating long-form content, such as novels, technical reports, and screenplays~\cite{wang2024autosurvey}.
However, compared to short-text generation, producing coherent and strictly length-constrained long documents remains a significant challenge~\cite{wu2025writingbenchcomprehensivebenchmarkgenerative,yang2025uncle,bailongwriter}.

Recent research has focused on enhancing LLMs with deep reasoning capabilities~\cite{shao2024deepseekmath,wang-etal-2025-lets}. While methods like DeepSeek-R1~\cite{guo2025deepseek} demonstrate impressive logic in mathematics and coding, their potential in open-ended long-form writing is less understood. In the writing domain, reasoning typically takes the form of static planning, where the model generates a comprehensive outline only at the initial stage~\cite{gurung2025learning,wang2025generating,wu2025longwriter}. To evaluate the limits of this paradigm, we conducted a systematic preliminary study (Section~\ref{sec:motivation}). Our experiments reveal a critical phenomenon that we term \textit{Length Collapse}: LLMs exhibit a sharp performance degradation as content length exceeds 2,000 words. 
We attribute this to the diminishing guidance of initial plans over long horizons, leading to a loss of narrative constraints. These findings suggest that one-shot reasoning is insufficient; effective long-form writing requires a mechanism to continuously update plans and reflect on progress.

To address this, we propose the \textbf{Interleaved Structural Chain-of-Thought (\texttt{IS-CoT})} framework. As illustrated in Figure~\ref{fig:paradigm}, unlike static outlining methods, \texttt{IS-CoT} enforces a dynamic, recursive workflow comprising \textit{global planning, local planning, content generation, and reflection}. This ensures that local execution remains aligned with global goals throughout the generation process. We implement this via a three-stage pipeline to construct a high-quality dataset of 5,000 samples with explicit interleaved reasoning traces, enabling the model to internalize this dynamic planning process.

By fine-tuning Qwen3-8B on this high-quality corpus, we develop IS-Writer-8B. 
Evaluations on LongBench-Write and WritingBench demonstrate that our model achieves state-of-the-art performance with average scores of 88.25 and 8.60, respectively. 
Our model not only surpasses the proprietary Gemini-2.5-Flash (+4.58) but also significantly outperforms larger open-source LLMs (e.g., DeepSeek-V3.2-671B) and other writing-enhanced LLMs (+10.28) on LongBench-Write.
Notably, IS-Writer demonstrates consistent generalization across different length ranges without overfitting. 
This robustness is particularly strong in the challenging ultra-long [4k, 20k) range, where our method effectively mitigates generation collapse compared to baselines.
Furthermore, our ablation studies verify the critical contributions of the \textit{interleaved planning} and \textit{reflection} modules, while case studies further highlight IS-Writer's precise controllability. 
In short, our contributions are:

\begin{itemize}[leftmargin=*]
\setlength{\itemsep}{2pt} 
\setlength{\parskip}{1pt}  
    \item We empirically identify \textbf{Length Collapse} in long-form generation, demonstrating that static planning is insufficient for ultra-long horizons.
    \item We propose \textbf{\texttt{IS-CoT}} and a dataset with explicit \texttt{Plan-Write-Reflect} traces, enabling process supervision for dynamic structural planning.
    \item We train \textbf{IS-Writer-8B}, which achieves SOTA performance on long-form generation, demonstrating that a smaller model with dynamic reasoning can outperform much larger baselines. 
\end{itemize}

\section{Motivation and Preliminary Study}
\label{sec:motivation}

To investigate the limitations of LLMs in long-form generation, we conduct two preliminary studies: quantifying performance degradation across target lengths and evaluating different planning strategies.

\subsection{\textit{Length Collapse} in Long-Form Generation}
\label{subsec:study1}
\begin{figure}[h] 
\centering
\includegraphics[scale=0.29]{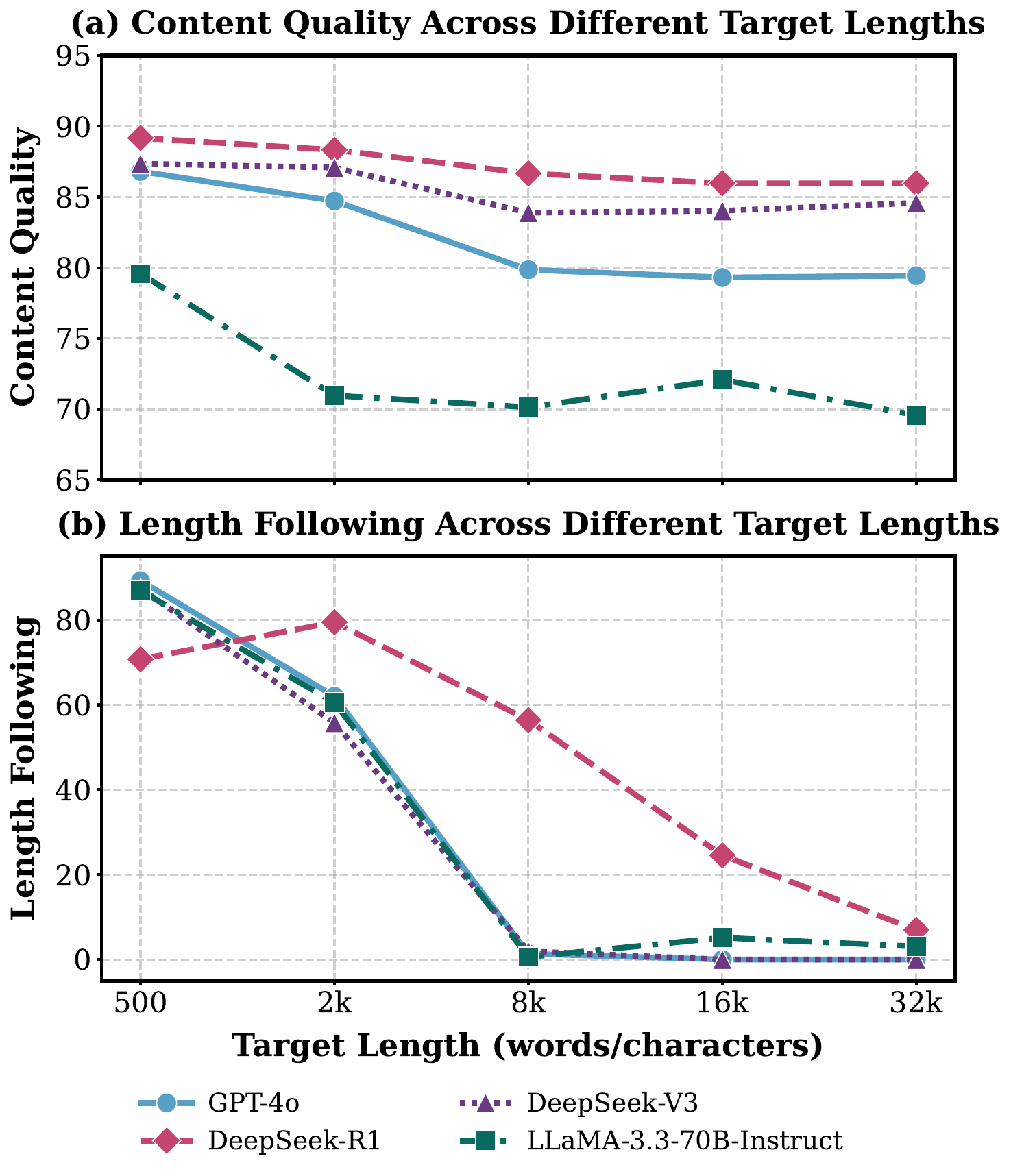}
\caption{Comparison of quality and length-following scores for LLMs across increasing target generation lengths, with quality scores evaluated by GPT-4o-mini.}
\label{fig:length}
\end{figure}

\begin{figure*}[h] 
\centering
\includegraphics[scale=0.27]{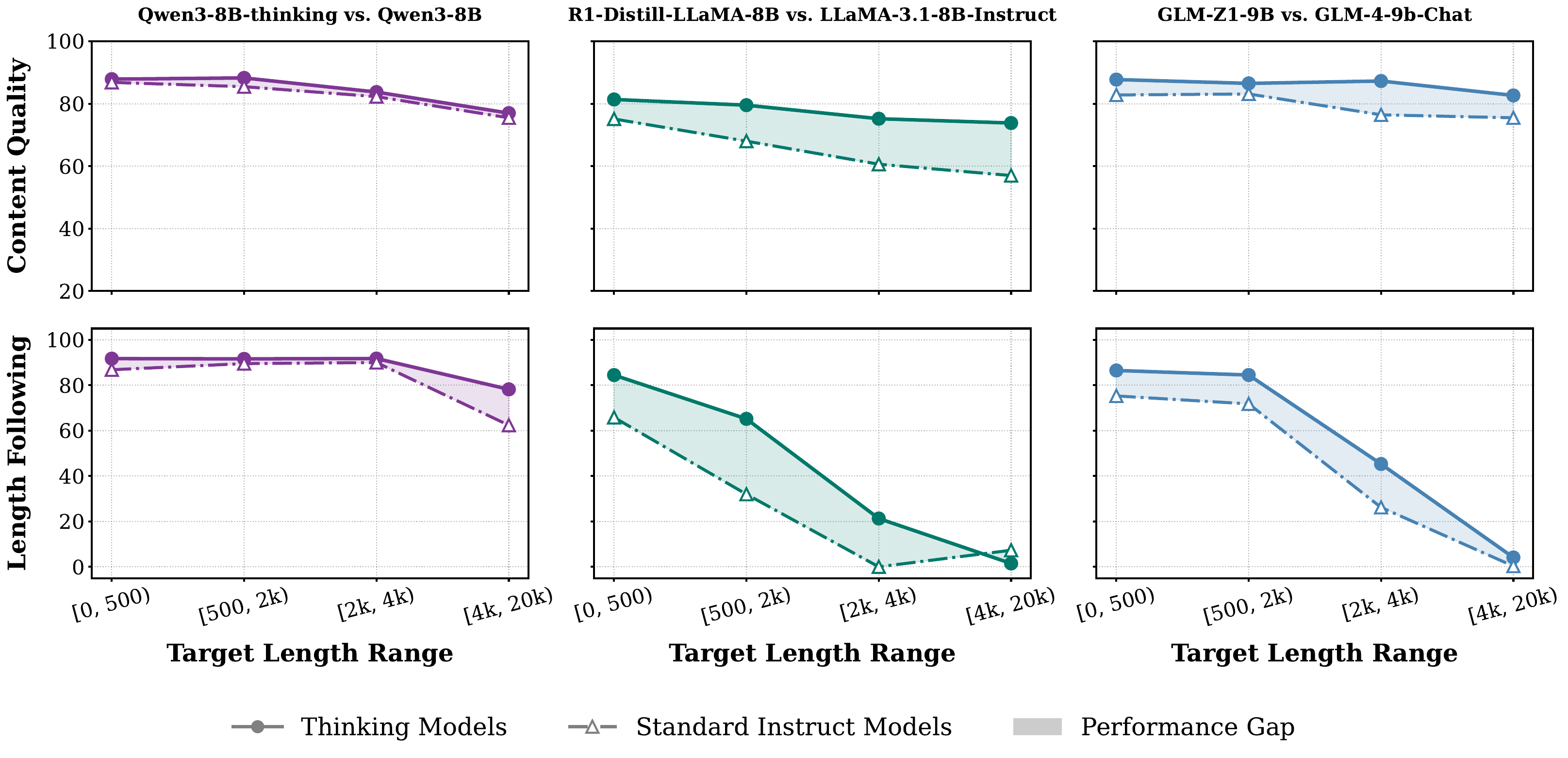}
\caption{Performance comparison between reasoning-enhanced models (``Thinking'') and standard instruct/chat models on LongBench-Write. Despite their superiority, ``Thinking'' models still degrade as length increases.}  
\label{fig:conpare}
\end{figure*}

While LLMs demonstrate strong performance in short text generation, their stability often decreases when generating very long content. In this study, we aim to measure this decline and explore the potential of reasoning capabilities.

\paragraph{Experimental Setup.}
We evaluate four representative models: GPT-4o~\cite{hurst2024gpt}, LLaMA-3.3-70B-Instruct~\cite{dubey2024llama}, DeepSeek-V3~\cite{liu2024deepseek}, and the reasoning-enhanced DeepSeek-R1~\cite{guo2025deepseek}. We constructed a dataset of 30 bilingual prompts (15 English/15 Chinese) across five target lengths (500 to 32,000 words), yielding 150 test cases per model evaluated via two metrics:

\begin{itemize}[leftmargin=*]
\setlength{\itemsep}{0pt}
\item \textbf{Quality Score (0-100):} Following previous work~\cite{bailongwriter}, we adopt LLM-as-a-judge to evaluate content quality on six dimensions: Relevance, Accuracy, Coherence, Clarity, Breadth and Depth, and Reading Experience.

\item \textbf{Length Following Score (0-100):} We calculate the length score $S_l$ using a piecewise linear function based on target length $l$ and output length $l'$. The score is 100 for exact matches and decays linearly to 0 at $l/3$ and $4l$, formulated as:
\begin{equation}
\resizebox{0.85\linewidth}{!}{$
    S_l = \begin{cases} 
    100 \cdot \max(0, 1 - (l'/l - 1)/3) & \text{if } l' > l \\
    100 \cdot \max(0, 1 - (l/l' - 1)/2) & \text{if } l' \leq l
    \end{cases}
$}
\end{equation}
\end{itemize}

\paragraph{Results.}
As shown in Figure~\ref{fig:length}, with the increasing of required generation length, we observe two typical phenomena: (1) quality degrades; and (2) generation length collapse.
Specifically, figure~\ref{fig:length}~(a) shows an inverse correlation between generation quality and target length, with the average Quality Score dropping by 5.83 across the [500, 32k] range. Notably, LLaMA-3.3 exhibits a more significant decline ($79.58 \to 69.58$) than GPT-4o ($86.81 \to 79.44$), highlighting the challenge of maintaining coherence in long-form generation.
Figure~\ref{fig:length}~(b) indicates that length following degrades much more sharply than quality. The average score drops by 19.13 points from 500 to 2,000 words, followed by a critical collapse between 2,000 and 8,000 words, where the score falls to 15.07 ($\Delta \approx -68.5$).
Models demonstrate complete failure at longer targets (16k, 32k) as average scores fall below 10 consistently.

\paragraph{Insight.}
DeepSeek-R1 outperforms all baselines on both metrics, exhibiting a significant advantage in length following for longer targets. This suggests the effectiveness of its internal ``thinking'' process in long-form generation. This observation motivates our core hypothesis: continuous planning throughout generation, rather than solely at the beginning, mitigates length and quality degradation. To explore this, we first examine the limitations of current planning methods in the following section.

\subsection{Benefits and Limitations of ``Thinking''}
\label{subsec:study2}

DeepSeek-R1's strong performance observed in Study 1 suggests that explicit reasoning capabilities (or ``thinking'' processes) are crucial for long-form generation. To confirm and verify this impact, we conducted a controlled comparison between models equipped with reasoning capabilities and their standard instruction-tuned versions.

\paragraph{Experimental Setup.}
We evaluated three pairs of models on the \textit{LongBench-Write} benchmark. Each pair consists of a standard Instruct/Chat model and its reasoning-enhanced variant:
(1) Qwen3-8B-Thinking vs. Qwen3-8B~\cite{yang2025qwen3};
(2) R1-Distill-LLaMA-8B~\cite{guo2025deepseek} vs. LLaMA-3.1-8B-Instruct~\cite{dubey2024llama};
(3) GLM-Z1-9B vs. GLM-4-9B-Chat~\cite{glm2024chatglm}.
Consistent with Study 1, we report both the Quality Score and Length Following Score to assess performance across varying target lengths.

\begin{figure*}[h]
\centering
\includegraphics[scale=0.22]{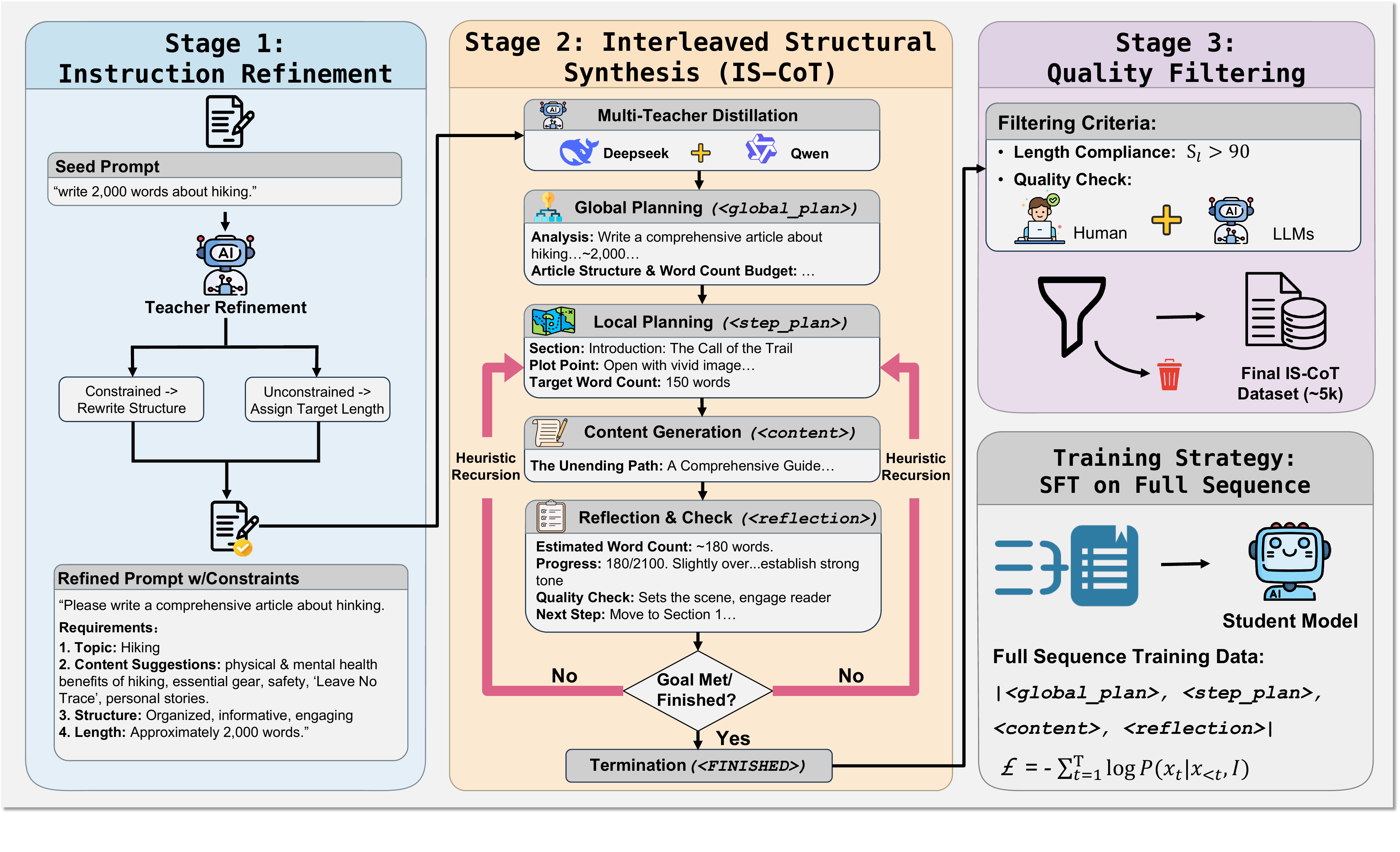}
\caption{The overall framework of \texttt{IS-CoT}. We construct the \texttt{IS-CoT} Dataset through a three-stage pipeline: (1) Instruction Refinement, (2) Interleaved Structural Synthesis via a Multi-Teacher Distillation Framework, and (3) Quality Filtering. Finally, we perform SFT to internalize these reasoning capabilities.}
\label{fig:main}
\end{figure*}

\paragraph{Results.}
Figure~\ref{fig:conpare} demonstrate a significant advantage for the ``think-before-you-write'' paradigm, yet reveal the limitations of static planning at long horizons.
As shown in Figure~\ref{fig:conpare}, the Thinking variants consistently outperformed their standard counterparts in both generation quality and length compliance across all three model families. This gap was particularly pronounced in the LLaMA series (average +14.57) and the GLM series (+9.15), indicating that initial planning leads to better performance than direct generation.
However, despite outperforming baselines, the efficacy of this initial reasoning diminishes as the target length increases.
For instance, the best-performing Qwen3-Thinking experienced a decrease in both Quality Score~(-10.89) and Length Following Score~(-13.53), with even more severe declines observed in the R1-Distill and GLM-Z1 models.
This suggests that while initial reasoning provides a strong starting point, it is insufficient to maintain coherence and constraint adherence throughout the generation of 2,000+ words, highlighting a critical issue.

\paragraph{Takeaway: The necessity of interleaved guidance.}
These findings highlight a critical limitation in current reasoning paradigms: the reliance on \textit{one-shot planning}.
In standard reasoning models, ``Thinking'' occurs entirely at the beginning. However, the model tends to ``forget'' this initial plan during extensive generation.
We argue that a static initial plan is insufficient for effective long-form generation. Instead, the model requires a mechanism to continuously guide and adjust its writing trajectory. This motivates our proposed Interleaved Structural CoT framework, which introduces a \texttt{Plan-Write-Reflect} cycle to support high-quality generation, as detailed in Section~\ref{sec:methodology}.

\section{Methodology}
\label{sec:methodology}

This section details our methodology for constructing the \textbf{Interleaved Structural CoT (\texttt{IS-CoT})} dataset and the subsequent training strategy. 
Embedding a \texttt{Plan-Write-Reflect} cycle into the training data enables the model to achieve fine-grained control over structure and length in long-form generation.
As illustrated in Figure~\ref{fig:main}, our data construction pipeline proceeds in three sequential stages: Instruction Refinement, Interleaved Structural Synthesis, and Quality Filtering.

\subsection{Stage I: Instruction Refinement}
Developing models capable of controllable long-horizon planning demands a standardized set of challenging writing tasks.
We start by sampling 6,000 seed prompts from the \textit{DeepWriting}~\cite{wang2025reverse} dataset, balanced between English and Chinese. This subset consists of 3,000 prompts with explicit length constraints and 3,000 without.
To ensure high-quality supervision, we utilize DeepSeek-V3.2 to enhance these prompts. The process specifically addresses two scenarios:
\begin{itemize}[leftmargin=*]
\setlength{\itemsep}{-1pt} 
\setlength{\parskip}{1pt}  
    \item \textbf{Constrained Prompts:} It enhances clarity and structure by rephrasing the instructions.
    \item \textbf{Unconstrained Prompts:} It assigns a challenging target length, with a preference for long-form tasks to facilitate long-context training.
\end{itemize}
The outcome is a set of 6,000 clear writing instructions, each with an explicit length target.

\subsection{Stage II: Interleaved Structural Synthesis}
\label{sec:synthesis}

Standard CoT typically generates reasoning only at the beginning, which often leads to guidance decay in long-form generation. To address this, we design a recursive generation workflow that interleaves planning, writing, and reflection.

\paragraph{Multi-Teacher Distillation Framework.}
We employ DeepSeek-V3.2 and Qwen3-235B-A22B-Instruct as teacher models for data synthesis. This dual-source strategy exposes the student model to diverse writing styles and distinct reasoning patterns, effectively preventing it from overfitting to the specific biases of a single teacher.

\paragraph{The Generation Workflow.}
We formulate long-form writing as an iterative decision-making process. To enable explicit planning and reflection, the generation is structured into four components defined by specific tokens:
\begin{enumerate}[leftmargin=*]
\setlength{\itemsep}{0pt} 
    \item \textbf{Global Planning (\texttt{<global\_plan>})}: Constructs a high-level outline based on the prompt. It decomposes the task into sections and assigns a length budget to each, ensuring the global structure satisfies the total length requirement.
    \item \textbf{Local Planning (\texttt{<step\_plan>})}: Designs the trajectory for the current segment. This includes outlining points, specifying constraints, and defining the length for the current segment.
    \item \textbf{Content Generation (\texttt{<content>})}: 
    Generates content based on the local plan, strictly adhering to the required format (e.g., report structure).
    \item \textbf{Reflection (\texttt{<reflection>})}: Performs a reflection after each segment. The model calculates length progress, verifies logical coherence, and adjusts the subsequent plan if necessary.
    
\end{enumerate}

\paragraph{Heuristic-Guided Recursive Generation.}
To ensure strict adherence to the interleaved format and length constraints, we implement a heuristic control loop during synthesis. We continuously inspect the output for the \texttt{<reflection>} token to immediately evaluate the current completion status:
\begin{itemize}[leftmargin=*]
\setlength{\itemsep}{0pt}
    \item \textbf{Continuation:} If the accumulated length is below the target, the system forces the generation of a \texttt{<step\_plan>} token, prompting the model to initiate planning for the subsequent section.
    \item \textbf{Termination:} Conversely, if the content is complete and satisfies all constraints, the loop concludes with a \texttt{<FINISHED>} token.
\end{itemize}
This active intervention ensures structural integrity and effectively mitigates the premature termination commonly observed in standard LLMs generation.

\subsection{Stage III: Quality Filtering}
Following data synthesis, we apply a rigorous filtering pipeline to guarantee both generation quality and length compliance. Quality assessment combines LLM-as-a-Judge with manual verification. For length compliance, we utilize the metric $S_{l}$ from Section~\ref{subsec:study1}, retaining only samples where $S_{l} > 90$. This yields the final \texttt{IS-CoT} dataset of approximately 5,000 high-quality samples.

\subsection{Training Strategy}
We train our student models via Supervised Fine-Tuning (SFT) on the \texttt{IS-CoT} dataset. Unlike standard methods that remove intermediate rationale, we train on the complete sequence including \texttt{<global\_plan>}, \texttt{<step\_plan>}, and \texttt{<reflection>} tokens. The objective function is the standard autoregressive negative log-likelihood:
\begin{equation}
    \resizebox{0.65\hsize}{!}{
        $ \mathcal{L} = - \sum_{t=1}^{T} \log P(x_t | x_{<t}, \mathcal{I}), $
    }
\end{equation}
where $x$ denotes the unified sequence of reasoning and content tokens, and $\mathcal{I}$ is the instruction. This explicitly forces the model to learn planning and reflection as core components of a coherent writing process, rather than as additional tasks.

\section{Experiments}

\definecolor{lightblue}{RGB}{235, 245, 255}   
\definecolor{mediumblue}{RGB}{210, 235, 255}  
\definecolor{deepblue}{RGB}{190, 215, 255}    

\begin{table*}[t]
\centering
\small 
\resizebox{1.0\textwidth}{!}{

\begin{tabular}{l | c | cc | c | cc | cccccc} 
\toprule
\multirow{2}{*}{\textbf{Models}} & \multicolumn{4}{c|}{\textbf{LongBench-Write}} & \multicolumn{8}{c}{\textbf{WritingBench}} \\ 
\cmidrule(lr){2-5} \cmidrule(lr){6-13} 
 & \textbf{Avg.} & $\boldsymbol{S_l}$ & $\boldsymbol{S_q}$ & \textbf{[4k, 20k)} &\textbf{L\_R} & \textbf{L\_C} & \textbf{D1} & \textbf{D2} & \textbf{D3} & \textbf{D4} & \textbf{D5} & \textbf{D6} \\ 

\midrule
\multicolumn{13}{l}{\cellcolor{lightblue}\textit{Proprietary LLMs}} \\ 
\midrule
GPT-4o & 71.61 & 57.29 & 85.94 &41.09 &7.35 & 7.16 & 7.39 & 6.86 & 7.03 & 7.45 & 7.48 & 7.70 \\
GPT-4o-mini & 72.22 & 59.45 & 84.98 &43.60 & 7.47 & 7.41 & 7.41 & 7.01 & 7.18 & 7.56 & 7.73 & 7.78 \\
Claude-3.5-sonnet & 72.90 & 60.17 & 85.63& 45.74 & 7.34 & 7.08 & 7.23 & 7.19 & 6.64 & 7.41 & 7.71 & 7.80 \\
Gemini-2.5-flash & 83.67 & 79.35 & 87.99 & \underline{82.11} & 8.47 & \underline{8.49} & 8.44 & 8.41 & 8.48 & 8.55 & 8.49 & 8.43 \\

\midrule
\multicolumn{13}{l}{\cellcolor{lightblue}\textit{Open-source LLMs}} \\
\midrule
DeepSeek-R1-671B & 81.87 & 75.31 & 88.44 &70.81& \underline{8.48} & 8.23 & \textbf{8.58} & 8.52 & 8.57 & 8.50 & 8.46 & 8.27 \\
DeepSeek-V3-671B & 75.10 & 62.76 & 87.43 &49.92& 8.05 & 7.57 & 7.93 & 7.77 & 8.11 & 8.10 & 8.11 & 8.10 \\
DeepSeek-V3.2-671B & 85.17 & 80.24 & \textbf{90.10} &71.03& \textbf{8.60} & \textbf{8.59} & 8.53 & \underline{8.55} & \underline{8.59} & \textbf{8.73} & \textbf{8.63} & \textbf{8.50} \\
Qwen3-235B-A22B-Instruct & \underline{87.15} & \underline{85.07} & \underline{89.24} &71.91& 8.45 & 8.45 & \textbf{8.58} & 8.43 & 8.40 & 8.60 & 8.44 & 8.26 \\
LLaMA3.3-70B-Instruct & 66.45 & 55.58 & 77.33 & 35.45& 6.58 & 6.40 & 6.52 & 6.10 & 6.26 & 6.43 & 7.13 & 7.01 \\
\midrule

\multicolumn{13}{l}{\cellcolor{lightblue}\textit{Capability-enhanced LLMs}} \\
\midrule
Suri-I-ORPO-7B & 44.25 & 42.42 & 46.08 & 29.67& 2.24 & 2.05 & 2.45 & 2.70 & 1.83 & 2.00 & 2.41 & 2.43 \\
LongWriter-8B & 77.97 & 77.25 & 78.68 & 75.37 & 4.44 & 4.14 & 4.50 & 3.66 & 4.12 & 4.13 & 5.24 & 5.00 \\
Writing-Model-Qwen-7B & 70.24 & 58.77 & 81.70 &39.77 & 8.35 & 8.39 & \underline{8.57} & 8.26 & 8.38 & 8.21 & 8.49 & 8.30 \\
Deepwriter-8B & 57.94 & 41.98 & 73.90 & 37.04 & 5.81 & 5.44 & 6.29 & 4.69 & 5.23 & 5.79 & 6.23 & 6.46 \\
\textbf{IS-Writer-8B~(ours)} & \textbf{88.25} & \textbf{88.31} & 88.19 & \textbf{85.46} & \textbf{8.60} & 8.44 & 8.56 & \textbf{8.61} & \textbf{8.60} & \underline{8.68} & \underline{8.60} & \underline{8.47} \\

\bottomrule
\end{tabular}}
\caption{Main performance comparison on LongBench-Write (scale: 0-100; see Appendix~\ref{sec:appendix_longbench} for details) and WritingBench (scale: 1-10). The six domains evaluated in WritingBench include: (D1) Academic \& Engineering, (D2) Finance \& Business, (D3) Politics \& Law, (D4) Literature \& Art, (D5) Education, and (D6) Advertising \& Marketing. The \textbf{best} and \underline{second-best} results are highlighted in bold and underlined, respectively.}
\label{tab:my_table}
\end{table*}

This section presents the experimental setup, comparative results against advanced baselines, ablation study and case study for our \textbf{IS-Writer-8B}.
\subsection{Experimental Setup}

\paragraph{Implementation Detail.}
IS-Writer-8B is fine-tuned from Qwen3-8B on our \texttt{IS-CoT} dataset. Training utilizes DeepSpeed ZeRO-3 across 64 NVIDIA H800 GPUs with a global batch size of 64 and a learning rate of $2 \times 10^{-5}$. To capture long-range dependencies, we extend the context window to 32,768 tokens and train for 5 epochs.

\paragraph{Baseline.}
We compare IS-Writer-8B against three categories of strong competitors:
\begin{itemize}[leftmargin=*]
\setlength{\itemsep}{0pt} 
\setlength{\parskip}{1pt} 
    \item \textbf{Proprietary LLMs:} Representative closed-source models, including GPT-4o, GPT-4o-mini, Claude-3.5-Sonnet~\cite{anthropic_claude35_sonnet_2024}, and Gemini-2.5-Flash~\cite{comanici2025gemini}.
    \item \textbf{Open-Source LLMs:} Large-scale open-weight models, including DeepSeek-R1, DeepSeek-V3, DeepSeek-V3.2~\cite{deepseekai2025deepseekv32pushingfrontieropen}, Qwen3-235B-Instruct~\cite{yang2025qwen3}, and LLaMA3.3-70B-Instruct~\cite{dubey2024llama}.
    \item \textbf{Capability-enhanced LLMs:} Models specialized in long-context generation, including Suri-I-ORPO~\cite{pham-etal-2024-suri}, LongWriter-8B~\cite{bailongwriter}, Writing-Model-Qwen-7B~\cite{wu2025writingbenchcomprehensivebenchmarkgenerative}, and DeepWriter-8B~\cite{wang2025reverse}.
\end{itemize}

\paragraph{Benchmark and Evaluation.}
We employ \textit{LongBench-Write}~\cite{bailongwriter} to assess comprehensive capabilities in ultra-long generation, measuring both length compliance and quality metrics. Additionally, we leverage the length-constrained subset of \textit{WritingBench}~\cite{wu2025writingbenchcomprehensivebenchmarkgenerative} and assess performance across six writing domains. Following official recommendations, we adopt the LLM-as-a-Judge with GPT-4o-mini to ensure fair and scalable evaluation.

\subsection{Main Results}

\begin{table*}[t]
    \centering
    \small
    \resizebox{\linewidth}{!}{
    \begin{tabular}{l|ccc|cc|cc|cc|cc}
    \toprule
     \multirow{2}{*}{\textbf{Models}} & \multicolumn{3}{c|}{\textbf{Overall}} & \multicolumn{2}{c|}{\textbf{[0, 500)}} & \multicolumn{2}{c|}{\textbf{[500, 2k)}} & \multicolumn{2}{c|}{\textbf{[2k, 4k)}} & \multicolumn{2}{c}{\textbf{[4k, 20k)}} \\
     \cmidrule(lr){2-4} \cmidrule(lr){5-6} \cmidrule(lr){7-8} \cmidrule(lr){9-10} \cmidrule(lr){11-12}
     & $\bar{S}$ & $S_l$ & $S_q$ & $S_l$ & $S_q$ & $S_l$ & $S_q$ & $S_l$ & $S_q$ & $S_l$ & $S_q$ \\
    \midrule
   \textbf{Qwen3-8B} & 83.52 & 83.23 & 83.82 & 86.82 & 86.07 & \textbf{89.53} & \textbf{89.05} & \textbf{90.00} & 86.67 & 62.36 & 69.67 \\ 
    \textbf{IS-Writer-8B} & \textbf{88.25}~(+4.73) & \textbf{88.31} & \textbf{88.19} & \textbf{89.60} & \textbf{91.67} & 88.56 & 87.98 & 87.68 & \textbf{88.13} & \textbf{86.75} & \textbf{84.17} \\
    \quad \emph{w/o Reflection} & 85.97~(+2.45) & 85.90 & 86.04 & 88.95 & 86.85 & 85.47 & 86.34 & 87.09 & 86.46 & 81.80 & 84.17 \\
    \quad \emph{w/o Interleaved} & 84.07~(+0.55) & 83.03 & 85.11 & 81.04 & 85.55 & 85.05 & 85.95 & 80.49 & 86.67 & 84.15 & 81.88 \\
    \bottomrule
    \end{tabular}
    }
    \caption{Ablation results on LongBench-Write. \textbf{IS-Writer-8B} represents the full model, while \emph{w/o Reflection} and \emph{w/o Interleaved} denote the variants without the reflection module and interleaved planning, respectively.}
    \label{tb:longbench_write_ablation}
\end{table*}

Table~\ref{tab:my_table} presents the comprehensive performance comparison. Our proposed \textbf{IS-Writer-8B} achieves state-of-the-art performance among writing-enhanced LLMs, and remarkably surpasses several proprietary LLMs and much larger open-source models across both benchmarks.

\paragraph{Superiority in Comprehensive Long-Form Generation.}
IS-Writer-8B achieves an average score of 88.25 on LongBench-Write, outperforming the best proprietary model Gemini-2.5-Flash (+4.58) and Qwen3-235B (+1.10).
Notably, in the challenging [4k, 20k) length range detailed in Table~\ref{tab:my_table}, IS-Writer demonstrates a substantial advantage, surpassing the best open-source LLM (+13.55) and writing-enhanced LLM (+10.09).
While some baselines show higher quality scores ($S_q$), they suffer in length compliance ($S_l$), often benefiting from the judge's bias toward shorter outputs. In contrast, IS-Writer-8B achieves the highest length score ($S_l$: 88.31) without compromising coherence.
Similarly, on WritingBench, our model matches the much larger DeepSeek-V3.2 (8.60) and achieves SOTA results in Finance (D2) and Politics (D3). This confirms that our Interleaved Structural CoT paradigm enhances general writing capabilities.

\paragraph{Surpassing Teacher Models via Dynamic Planning.}
A significant finding is that IS-Writer outperforms the teacher models (Qwen3-235B and DeepSeek-V3.2) used to synthesize its training data.
While teacher models possess vast knowledge, they struggle to organize it over ultra-long contexts in a single inference pass. By distilling the writing process into explicit steps (Plan-Write-Reflect) and utilizing a multi-teacher framework for data robustness, IS-Writer-8B effectively learns a dynamic planning logic that proves more robust than the unstructured generation of its teachers.

\paragraph{Robustness in Ultra-Long Length Compliance.}
\begin{figure}[h] 
\centering
\includegraphics[scale=0.30]{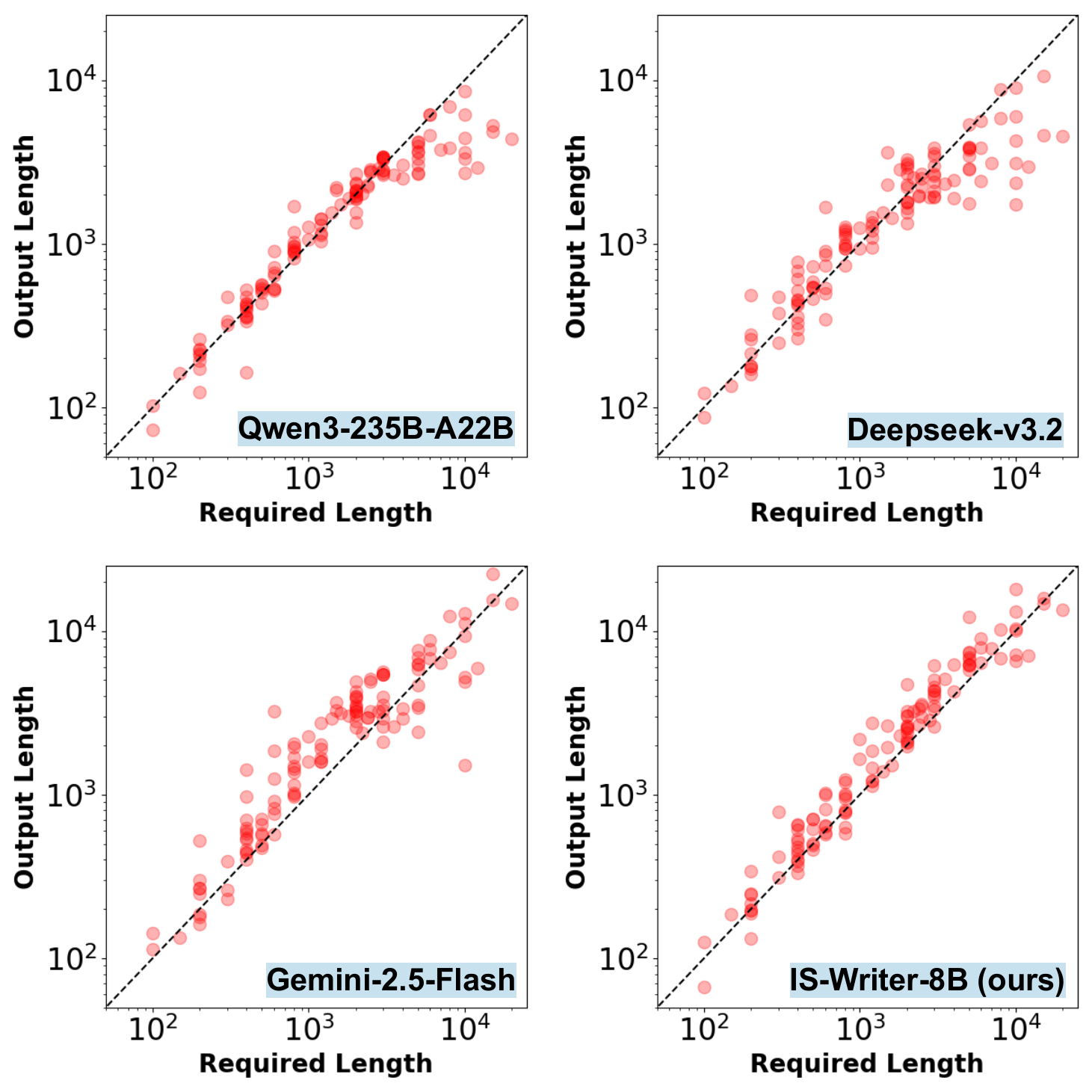}
\caption{Model response length w.r.t. instruction required length on LongBench-Write.}
\label{fig:lw_length}
\end{figure}

Figure~\ref{fig:lw_length} visualizes the length compliance in the range of [0, 20k). IS-Writer demonstrates strong robustness across various length constraints, especially in the challenging [4k, 20k) range. In this specific range, open-source LLMs exhibit a ``lazy'' behavior, tending to under-generate as requirements increase. While proprietary LLMs such as Gemini-2.5 perform better, they still show high variance. In contrast, IS-Writer consistently meets or slightly exceeds target lengths. This precision is attributed to the interleaved reasoning, where the model dynamically verifies its progress and adjusts next plans, effectively mitigating the ``early stoping'' common in long-form generation.

\subsection{Ablation Study}

To verify the contributions of core components within the \texttt{IS-CoT} framework, we conduct ablation studies on LongBench-Write, comparing the full model against variants with components removed.

\paragraph{Impact of Components.}
As shown in Table~\ref{tb:longbench_write_ablation}, IS-Writer-8B demonstrates a substantial overall improvement, outperforming the base Qwen3-8B by 4.73 points.
Removing the reflection module (\emph{w/o Reflection}) reduces the gain to +2.45, while further removing the local planning module (\emph{w/o Interleaved}) drops the improvement to just +0.55. 
This trend confirms that while global planning offers a basic direction, the combination of step-wise local planning and dynamic reflection is essential for maintaining high-quality long-form generation.

\paragraph{Robustness in Ultra-Long Contexts.}
The advantages of our method are most significant in the ultra-long [4k, 20k) range. 
Without explicit planning, Qwen3-8B performs poorly with an average score of 66.02. In contrast, IS-Writer-8B achieves an average score of 85.46~(+19.44). 
Notably, both ablation variants also score above 80 in this range, validating that explicit thinking processes, whether global or interleaved are critical for mitigating generation collapse in extended contexts.

\paragraph{Analysis of Short-Range Performance.}
We observe a small performance drop for IS-Writer in the [500, 2k) interval compared to the baseline. This is likely due to the data distribution in our training set because it contains fewer samples in this range. It may also relate to the bias of LLM judges toward standard responses for shorter texts. Nevertheless, considering the significant gains achieved in complex, long-form generation tasks, this slight decline is an acceptable trade-off for the model's specialized capability in handling extended outputs.

\subsection{Case Study}
\label{sec:case_study}

\begin{figure*}[h] 
\centering
\includegraphics[scale=0.31]{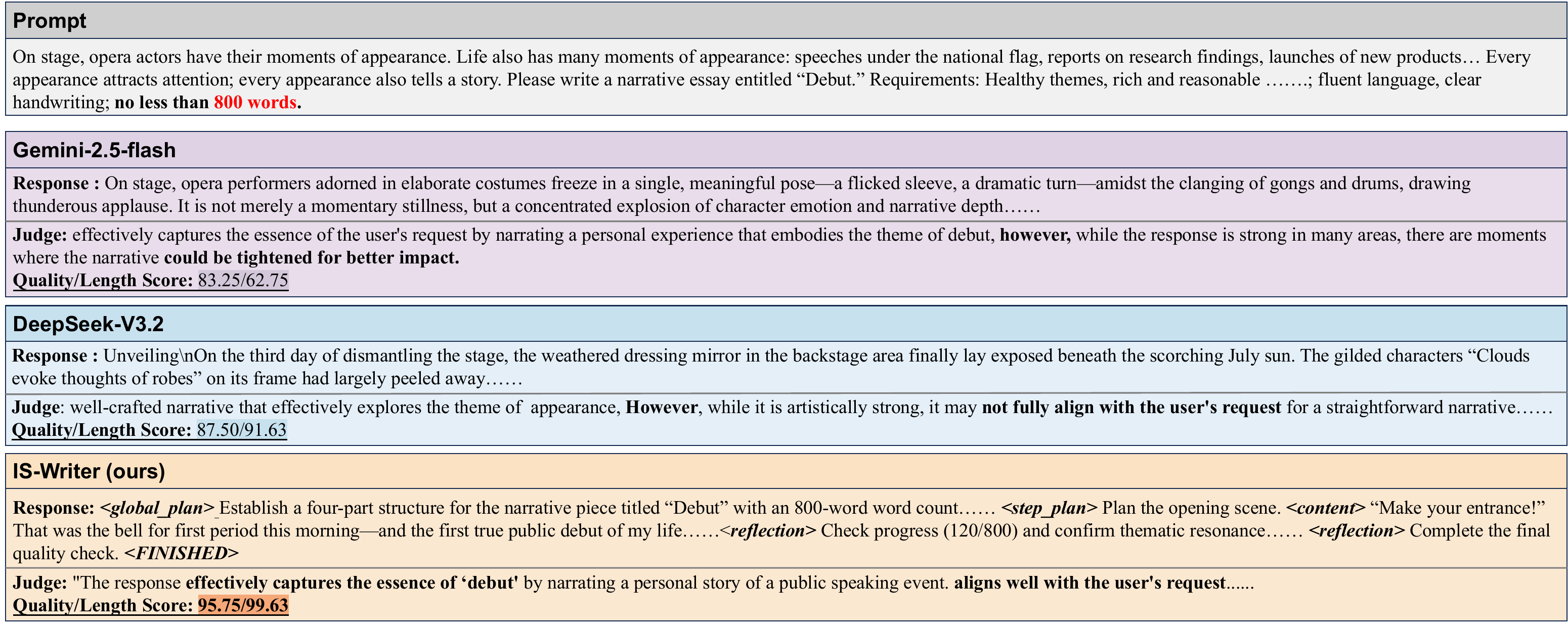}
\caption{Case study on a long-form narrative task from LongBench-Write, where IS-Writer uses interleaved reasoning, with additional case studies presented in Appendix~\ref{sec:appendix_case_studies}.} 
\label{fig:case}
\end{figure*}

We analyze the narrative task ``Debut'' (Figure~\ref{fig:case}) to demonstrate the efficacy of \texttt{IS-CoT}. While baselines like Gemini-2.5-Flash and DeepSeek-V3.2 produce coherent text, they often fail to meet specific length or narrative constraints due to the lack of explicit continuous planning. In contrast, \mbox{IS-Writer-8B} initiates with a \texttt{<global\_plan>} and interleaves generation with \texttt{<local\_plan>} and \texttt{<reflection>}. As shown in Figure~\ref{fig:case}, the model explicitly checks its progress (e.g., ``word count: 120/800'') in the reflection phase and adjusts the next writing plan accordingly. This dynamic mechanism ensures strict constraint adherence, yielding a length compliance score of 99.6. This demonstrates that controllable long-form generation requires a unified process of continuous planning and reflection rather than a large context window.

\section{Related Work}
\paragraph{Long-Form Generation and Planning.}
While Large Language Models (LLMs) excel in short-context tasks, they struggle to maintain coherence in open-ended long-form generation~\cite{ping-etal-2025-longdpo,lei2025writing,que2025pic}. Extended sequences often suffer from progressive quality decay~\cite{yang2025longfaith,yang2025uncle,liu2025verifact,he2025precise} and the ``lost-in-the-middle'' phenomenon~\cite{an2024make,baker2024lost}, leading to a negative length-quality correlation. To address these issues, hierarchical ``Plan-then-Write'' paradigms~\cite{bailongwriter,wu2025superwriter,zhao2025plan} have been proposed to decompose complex tasks into global outlines followed by sequential decoding. However, these frameworks typically rely on static planning that cannot adapt to content development during generation. In contrast, our approach replaces inflexible planning with an interleaved dynamic framework, enabling continuous alignment between plan and content.

\paragraph{Chain-of-Thought and Generation with Reasoning.} Chain-of-Thought (CoT) prompting has achieved significant success in logic-intensive domains like mathematics and coding~\cite{wei2022chain,shao2024deepseekmath,yang2025markov,xu2025redstar}. Recent studies advance this by distilling reasoning capabilities into smaller models~\cite{hsieh2023distilling,wangself,feng2024keypoint,dai2024beyond,wang2025reverse,zhuang2025unicott,yin2025marco}. Despite these advancements, explicit reasoning in creative long-form generation remains underexplored. Existing methods primarily confine reasoning to a pre-generation planning phase~\cite{liang2024integrating}, failing to leverage reasoning capabilities for intermediate adjustments. We bridge this gap by introducing an interleaved structural CoT framework that combines global planning, local reasoning, and reflection, allowing the model to dynamically refine its strategy based on the generated context.

\section{Conclusion}

In this work, we systematically evaluate current LLMs on long-form generation, revealing that even reasoning-enhanced models suffer from ``Length Collapse'' due to the diminishing effectiveness of static planning over extended horizons. To address this limitation, we present the \texttt{IS-CoT} framework, which introduces a dynamic ``Plan-Write-Reflect'' cycle to ensure continuous alignment with global constraints. By training on the \texttt{IS-CoT} dataset with explicit intermediate reasoning steps, our model learns to execute dynamic planning strategies through process supervision. Our experiments demonstrate that \textbf{IS-Writer-8B} achieves superior performance with remarkable data efficiency, surpassing significantly larger LLMs. This work underscores the necessity of dynamic guidance for long-context tasks and provides a robust framework for enhancing the controllability of LLMs.

\section*{Limitations}

Despite the promising performance of IS-Writer in controllable long-form generation, we acknowledge several limitations in our current work:

\paragraph{Dependency on Teacher Quality.}
Our training data relies on a multi-teacher synthesis pipeline. Although IS-Writer demonstrates performance that matches or even exceeds the teacher models in long-form generation tasks, the diversity and knowledge scope of the model are still influenced by the source supervision. We believe that integrating significantly stronger or a more diverse set of teachers in the future could further unlock the model's potential and generalization capabilities.
\paragraph{Inference Overhead.}
The core mechanism of \texttt{IS-CoT} involves generating interleaved ``Plan'' and ``Reflect'' tokens alongside the content. While this significantly enhances coherence and length compliance, it inevitably increases token consumption during inference. Compared to direct generation models, IS-Writer requires more computation time to produce the same amount of content. Future work could explore token-efficient mechanisms or reasoning approaches to mitigate this overhead.

\section*{Acknowledgments}
We want to thank all the anonymous reviewers for their valuable comments. This work was supported by the National Science Foundation of China (NSFC No. 62576232), the Postgraduate Research \& Practice Innovation Program of Jiangsu Province~(Grant No. KYCX25\_3467), Key Laboratory of Data Intelligence and Advanced Computing, Soochow University.

\bibliography{custom}
\appendix



\section{\texttt{IS-CoT} Dataset Statistics}
\label{sec:appendix_data_stats}

We provide detailed statistics of the 4,988 high-quality training samples in the \texttt{IS-CoT} dataset to demonstrate its diversity and coverage.

\subsection{Domain Distribution}

\begin{figure}[h]
    \centering
    \includegraphics[width=0.98\linewidth]{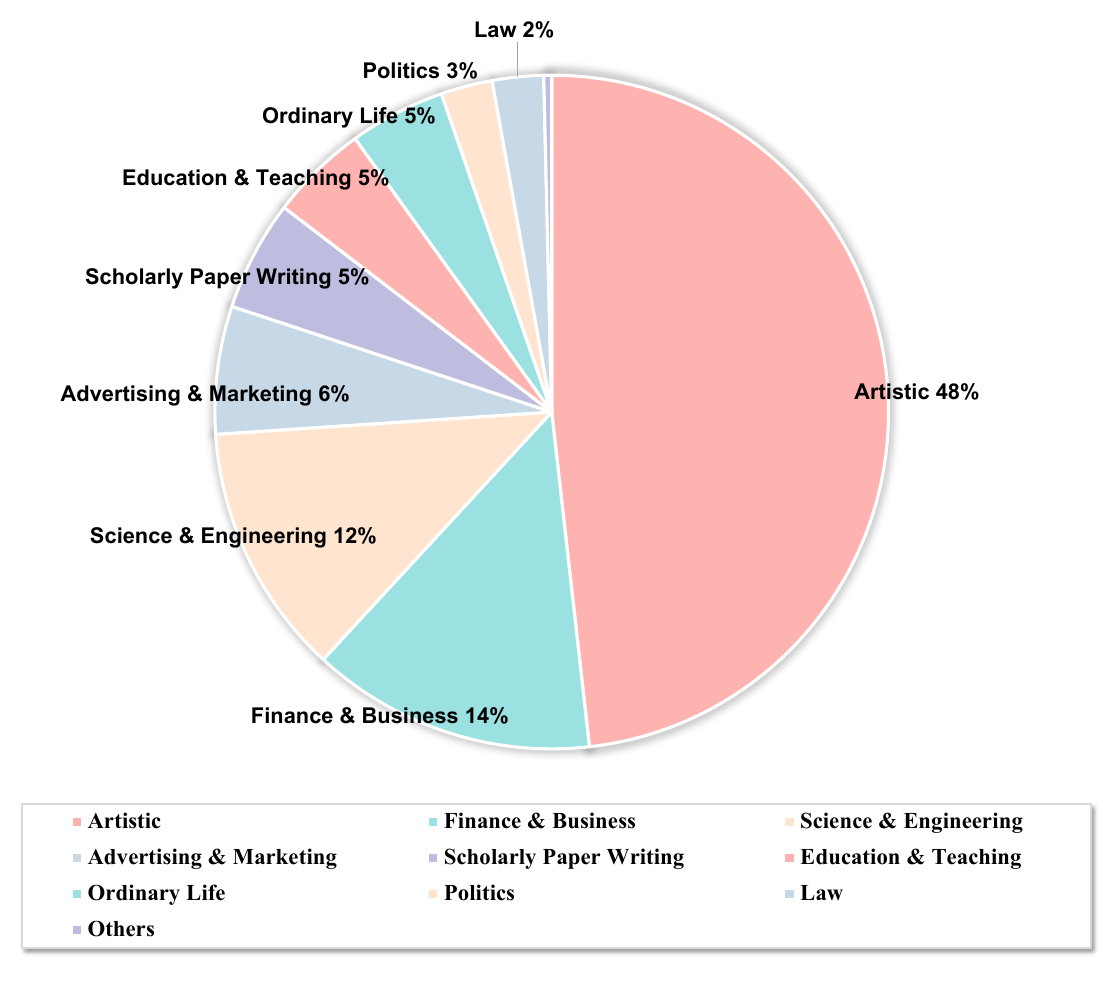}
    \caption{Domain distribution of the \texttt{IS-CoT} dataset. The diversity of topics ensures robust generalization.}
    \label{fig:domain_dist}
\end{figure}

As shown in Figure~\ref{fig:domain_dist}, the dataset covers a broad range of topics. ``Artistic'' writing (e.g., novels, scripts) makes up the largest part (48.4\%), helping the model learn narrative structures. Moreover, professional fields like ``Finance \& Business'' and ``Science \& Engineering'' are also well-represented. This mix ensures IS-Writer performs effectively for both creative stories and technical reports.

\subsection{Length Distribution}
\begin{figure*}[h]
    \centering
    \includegraphics[width=0.96\linewidth]{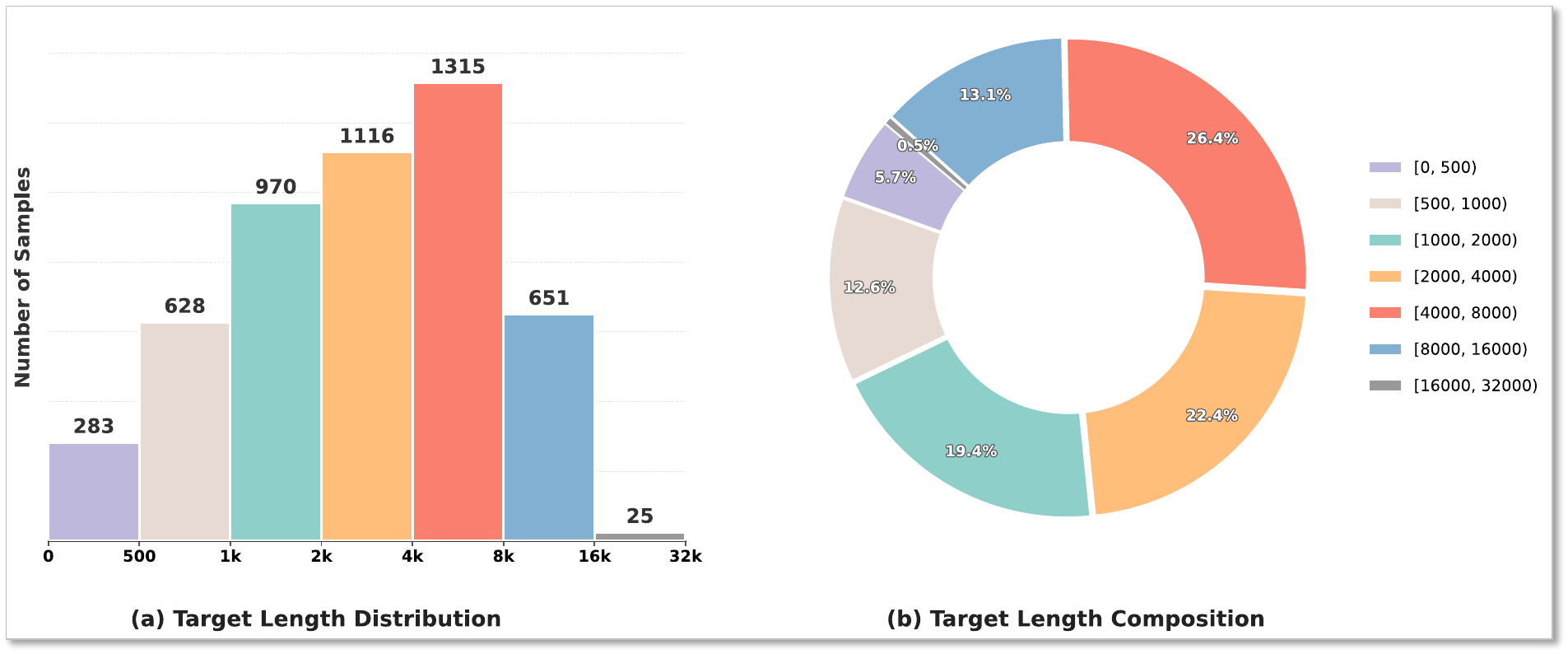}
    \caption{Target length distribution of the \texttt{IS-CoT} dataset. The dataset is concentrated on lengths over 2,000 words to support long-form generation.}
    \label{fig:length_dist}
\end{figure*}

To address the ``Length Collapse'' problem mentioned earlier, our dataset focuses on long-form generation. Figure~\ref{fig:length_dist} shows the length distribution. Over 60\% of the samples are longer than 2,000 words, with the [4,000, 8,000) range being the most common (26.36\%). We also include a significant number of very long samples (13.05\%) in the [8,000, 16,000) range. This helps the model learn the dynamic \textit{Plan-Write-Reflect} cycle for consistent and coherent long-document generation.

\begin{table*}[t]
    \centering
    \small
    \resizebox{\linewidth}{!}{
    \begin{tabular}{l|ccc|cc|cc|cc|cc}
    \toprule
     \multirow{2}{*}{\textbf{Models}} & \multicolumn{3}{c|}{\textbf{Overall}} & \multicolumn{2}{c|}{\textbf{[0, 500)}} & \multicolumn{2}{c|}{\textbf{[500, 2k)}} & \multicolumn{2}{c|}{\textbf{[2k, 4k)}} & \multicolumn{2}{c}{\textbf{[4k, 20k)}} \\
     \cmidrule(lr){2-4} \cmidrule(lr){5-6} \cmidrule(lr){7-8} \cmidrule(lr){9-10} \cmidrule(lr){11-12}
     & $\bar{S}$ & $S_l$ & $S_q$ & $S_l$ & $S_q$ & $S_l$ & $S_q$ & $S_l$ & $S_q$ & $S_l$ & $S_q$ \\
    \midrule
    \multicolumn{3}{l}{\emph{Proprietary LLMs}} \\
    \midrule
    \textbf{GPT-4o} & 71.61 & 57.29 & 85.94 & \textbf{94.39} & 89.58 & 79.28 & 86.43 & 22.24 & 83.75 & 0.00 & 82.17 \\
    \textbf{GPT-4o mini} & 75.56 & 65.84 & 85.28 & 93.65 & 87.76 & 86.70 & 85.76 & 52.08 & 84.58 & 5.37 & 81.83 \\
    \textbf{Claude 3.5 Sonnet} & 72.90 & 60.17 & 85.62 & 93.06 & 88.54 & 75.40 & 85.27 & 39.44 & 85.00 & 8.47 & 83.00 \\
    \textbf{Gemini-2.5-Flash} & 83.67 & 79.35 & 87.99 & 85.51 & 90.89 & 72.12 & 88.28 & 84.80 & 86.88 & 79.54 & 84.67 \\
    \midrule
    \multicolumn{3}{l}{\emph{Open-source LLMs}} \\
    \midrule
    \textbf{DeepSeek-R1} & 81.87 & 75.31 & 88.44 & 81.34 & 90.76 & 80.04 & 87.98 & 81.32 & 87.50 & 54.63 & 87.00 \\
    \textbf{DeepSeek-V3} & 75.10 & 62.76 & 87.43 & 90.65 & 90.63 & 81.18 & 88.57 & 35.55 & 85.83 & 17.17 & 82.67 \\
    \textbf{DeepSeek-V3.2} & 85.17 & 80.24 & \textbf{90.10} & 89.03 & 90.23 & 87.02 & \textbf{90.50} & 85.38 & \textbf{90.63} & 53.22 & \textbf{88.83} \\
    \textbf{Qwen3-235B-A22B-Instruct} & 87.15 & 85.07 & 89.24 & 92.09 & 89.06 & \textbf{93.12} & 89.44 & \textbf{93.73} & 90.00 & 55.31 & 88.50 \\
    \textbf{LLaMA3.3-70B-Instruct} & 66.45 & 55.58 & 77.33 & 87.13 & 83.85 & 78.06 & 77.23 & 24.88 & 76.46 & 1.08 & 69.83 \\
    \midrule
    \multicolumn{3}{l}{\emph{Capability-enhanced LLMs}} \\
    \midrule
    \textbf{Suri-I-ORPO (7B)} & 44.25 & 42.42 & 46.08 & 56.70 & 55.34 & 50.19 & 50.29 & 23.19 & 38.33 & 26.17 & 33.17 \\
    \textbf{LongWriter-8B} & 77.78 & 77.25 & 78.30 & 86.63 & 77.86 & 68.41 & 77.62 & 84.82 & 82.92 & 74.40 & 76.33 \\
    \textbf{Writing-Model-Qwen-7B} & 70.24 & 58.77 & 81.70 & 75.69 & 81.38 & 81.07 & 85.47 & 52.38 & 81.67 & 3.86 & 75.67 \\
    \textbf{DeepWriter-8B} & 61.26 & 47.93 & 74.58 & 76.13 & 80.99 & 62.77 & 75.29 & 21.79 & 72.50 & 7.24 & 66.83 \\
    \textbf{IS-Writer-8B} & \textbf{88.25} & \textbf{88.31} & 88.19 & 89.60 & \textbf{91.67} & 88.56 & 87.98 & 87.68 & 88.13 & \textbf{86.75} & 84.17 \\
    \bottomrule
    \end{tabular}
    }
    \caption{Detailed evaluation results on LongBench-Write across different length ranges. The best results are highlighted in \textbf{bold}, demonstrating that IS-Writer-8B achieves the highest overall average score.}
    \label{tb:longbench_write}
\end{table*}
\section{Detailed Results on LongBench-Write}
\label{sec:appendix_longbench}

We present detailed performance results of the models across different target length ranges on the LongBench-Write benchmark in Table~\ref{tb:longbench_write}.

\textbf{Comparison with Baselines.} 
IS-Writer-8B significantly outperforms all other capability-enhanced LLMs (e.g., LongWriter-8B, DeepWriter-8B) across every length. Furthermore, it generally surpasses leading proprietary models in overall scores, including GPT-4o and Gemini-2.5-Flash.

\textbf{Performance by Length Range.}
In the medium-length range (e.g., [500, 4k)), our model performs slightly below the best open-source models, such as DeepSeek-V3.2 and Qwen3-235B. We attribute this to our training data distribution, which is heavily weighted towards the [4k, 8k) range to focus on long-form generation. Nevertheless, our model remains competitive in these shorter length ranges.

Overall, IS-Writer-8B achieves the highest overall average score ($\bar{S} = 88.25$) among all evaluated models. This advantage is mainly due to its strong performance in the challenging ultra-long [4k, 20k) range. For instance, the length score of DeepSeek-V3.2 experiences a sharp decline to 53.22 in this specific range. While other models suffer from severe degradation (e.g., LLaMA3.3 dropping to 1.08 in length score), IS-Writer maintains a high length score ($S_l$) of 86.75.
This demonstrates that our approach effectively prevents the loss of control during long-form generation, thereby validating the effectiveness of the \texttt{IS-CoT} framework in maintaining precise controllability and coherence.

\begin{figure*}[h] 
\centering

\includegraphics[scale=0.31]{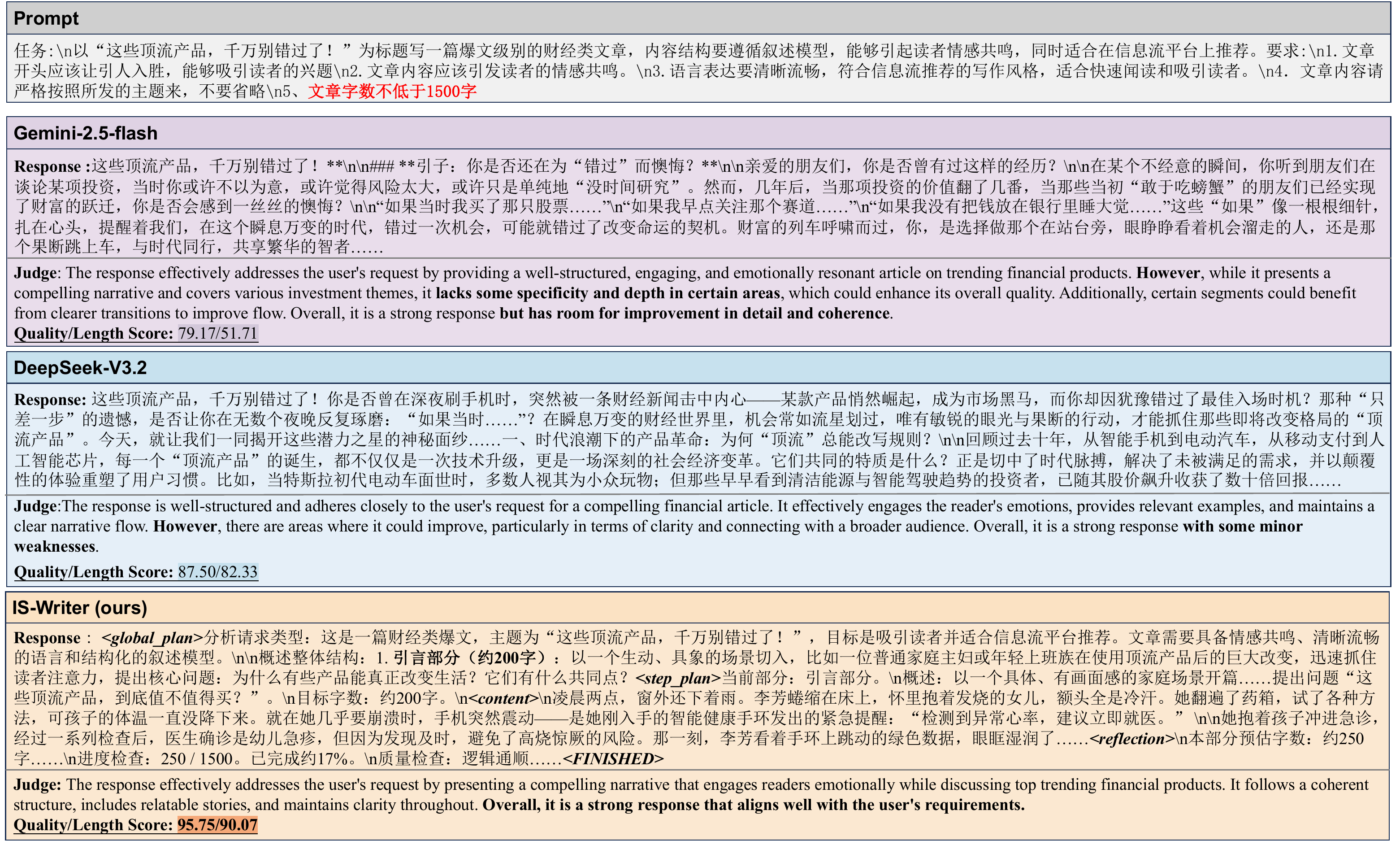}
\caption{Comparison on a Chinese creative writing task. Baselines struggle with 
length or depth, while IS-Writer generates a coherent article that strictly 
follows the $>$1500 word requirement.}
\label{fig:case_cn}

\vspace{1.5em}  

\includegraphics[scale=0.31]{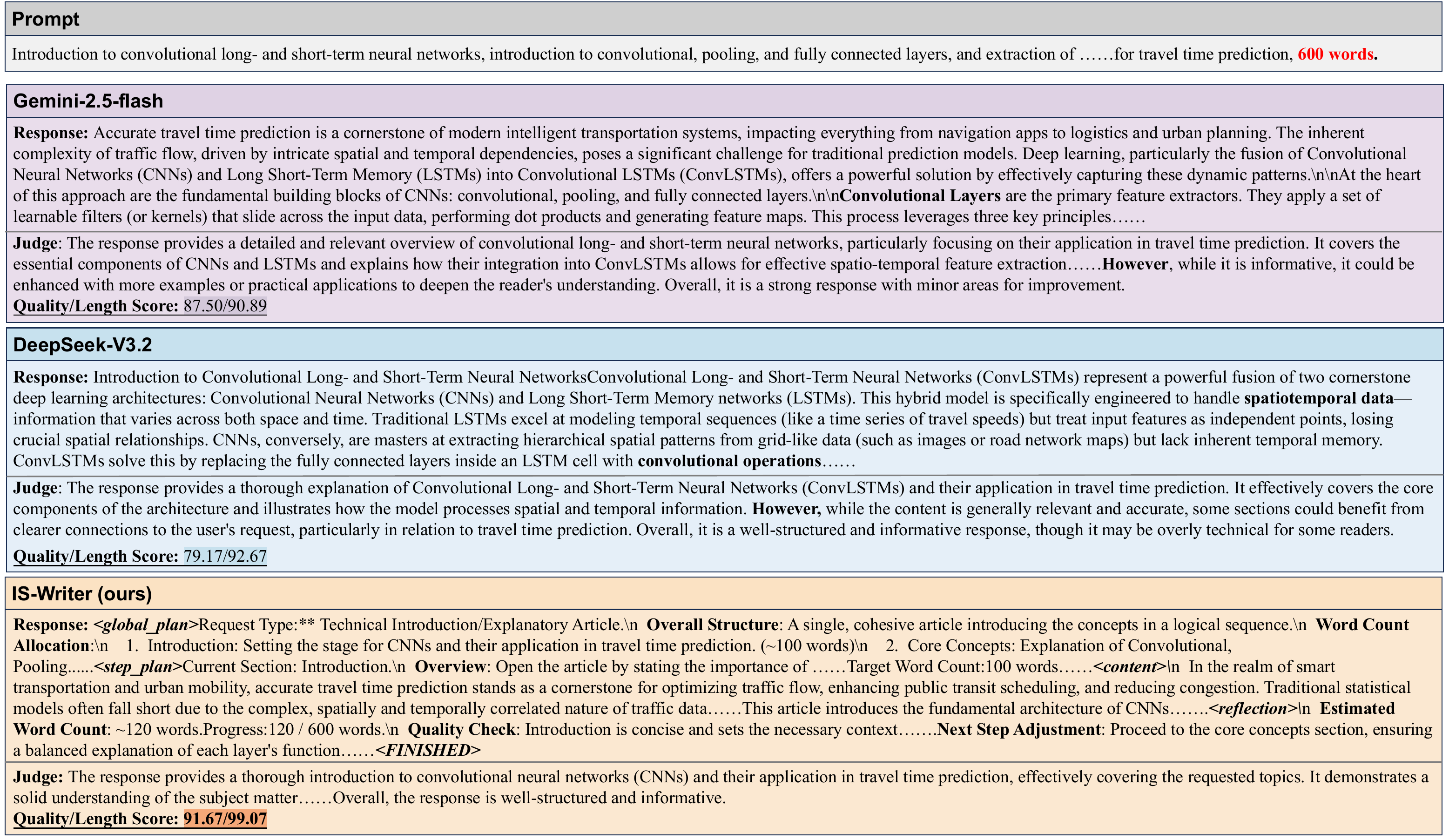}
\caption{Comparison on an English technical writing task. IS-Writer achieves 
better content quality and near-perfect length precision (Score: 99.07), 
outperforming baselines in both accuracy and control.}
\label{fig:case_en}

\end{figure*}

\section{Case Studies}
\label{sec:appendix_case_studies}

To demonstrate the performance of IS-Writer, we compare it against two strong baselines: Gemini-2.5-Flash and DeepSeek-V3.2. We select two distinct representative cases, covering Chinese creative writing and English technical writing, to evaluate generation quality and length control.

\subsection{Chinese Creative Writing}
The first case involves writing a popular financial article titled \textit{``Do not miss these top-tier products!''} The prompt requires a story-like structure to attract readers on a recommendation feed, with a minimum length of 1,500 words, creating a significant challenge for maintaining coherence. The comparison is shown in Figure~\ref{fig:case_cn}.
\textbf{Analysis:} 
Gemini-2.5-Flash fails to meet the length requirement, resulting in a low length score of 51.71. While the content is structured, it lacks the necessary depth. DeepSeek-V3.2 performs better (Quality: 87.50, Length: 82.33) but misses opportunities to connect with a broader audience. 
In contrast, IS-Writer achieves the highest scores in both aspects (Quality: 95.75, Length: 90.07). The evaluation shows that IS-Writer produces a ``compelling narrative'' with ``relatable stories'' while keeping the content clear. This proves that IS-Writer can handle long writing tasks effectively, maintaining engagement consistently throughout the entire document.

\subsection{English Technical Writing}
The second case requires a technical introduction to \textit{``Convolutional Long- and Short-Term Neural Networks (ConvLSTM)''} for travel time prediction, with a specific target length of about 600 words. The comparison is shown in Figure~\ref{fig:case_en}.
\textbf{Analysis:}
DeepSeek-V3.2 (Quality: 79.17) gives a detailed explanation but is considered ``overly technical,'' leading to a lower quality score. Gemini-2.5-Flash provides a strong response (Quality: 87.50) but lacks sufficient practical examples.
IS-Writer outperforms both baselines with a Quality score of 91.67 and a near-perfect Length score of 99.07. The evaluation highlights the response's ``solid understanding'' and ``well-structured'' format. The high length score confirms the robust effectiveness of the \texttt{IS-CoT} framework in precise control, ensuring the model follows the instructions precisely without generating unrelated content. This demonstrates that our method effectively optimizes content depth while respecting strict constraints.

\end{document}